\documentclass[10pt, a4paper]{article}

\usepackage{tipa}
\usepackage{microtype}
\usepackage{upgreek}
\usepackage{amsfonts}
\usepackage{amsmath}
\usepackage{hyperref}
\usepackage{cleveref}

\usepackage[]{lrec-coling2024} 

\def\numlangs{859}
\def\numtrans{1051}
\title{PrOnto: Language Model Evaluations for \numlangs{} Languages}

\name{Luke Gessler}
\address{Department of Computer Science \\ University of Colorado Boulder \\ \texttt{luke.gessler@colorado.edu} }

\abstract{
Evaluation datasets are critical resources for measuring the quality of pretrained language models.
However, due to the high cost of dataset annotation, these resources are scarce for most languages other than English, making it difficult to assess the quality of language models.
In this work, we present a new method for evaluation dataset construction which enables any language with a New Testament translation to receive a suite of evaluation datasets suitable for pretrained language model evaluation.
The method critically involves aligning verses with those in the New Testament portion of English OntoNotes, and then projecting annotations from English to the target language, with no manual annotation required.
We apply this method to \numtrans{} New Testament translations in \numlangs{} languages and make them publicly available.
Additionally, we conduct experiments which demonstrate the efficacy of our method for creating evaluation tasks which can assess language model quality. }

\begin{document}

\maketitleabstract

\def\CodeRepo{\url{https://github.com/lgessler/pronto}}

\def\wordtovec{\textsc{word2vec}}
\def\mbert{\textsc{mbert}}
\def\mbertva{\textsc{mbert-va}}
\def\m{$\upmu$\textsc{b}}
\def\mp{$\upmu$\textsc{b-p}}
\def\mt{$\upmu$\textsc{b-t}}
\def\mpt{$\upmu$\textsc{b-pt}}
\def\mptsla{$\upmu$\textsc{b-pt-sla}}
\def\mx{$\upmu$\textsc{b-x}}
\def\mxp{$\upmu$\textsc{b-xp}}
\def\mxt{$\upmu$\textsc{b-xt}}
\def\mxpt{$\upmu$\textsc{b-xpt}}
\def\mxptsla{$\upmu$\textsc{b-xpt-sla}}
\def\np{\textsc{-np}}
\def\pp{\textsc{-pp}}
\def\hqp{\textsc{-hqp}}
\def\bd{\textsc{-bd}}
\def\bm{\textsc{-bm}}

\section{Introduction}
Language models such as BERT \citep{devlin_bert_2019} and other Transformer-based \citep{vaswani_attention_2017} language models (TLMs) are notoriously difficult to understand.
Evaluation datasets such as SuperGLUE \citep{wang_superglue_2019}, BLiMP \cite{warstadt2020blimp}, and others have been essential resources for understanding and comparing different models' capabilities. 
By measuring two models' performance on a question-answering task, for example, we are able to make an assessment about the models' capabilities relative to each other.
Unfortunately, these evaluation tasks almost always require \textit{annotated} data produced by a human being, and these datasets are therefore very scarce except for the most well-resourced languages, especially English.
This scarcity of evaluation datasets has been a significant hindrance for research on TLMs for low-resource languages, as it is much harder to assess the quality and properties of models without them.

Here, we present PrOnto, a dataset consisting of {\bf pr}ojections of {\bf Onto}Notes' New Testament annotations into New Testament translations in \numlangs{} different languages.
OntoNotes \citeplanguageresource{hovy_ontonotes_2006} is a corpus with many annotation types covering a wide variety of phenomena in grammar and meaning.
A subset of the English portion of OntoNotes contains the Easy-to-Read Version (ERV) translation of the New Testament, complete with a segmentation of each sentence into the book, chapter, and verse of the Bible that it appeared in.
Using these verse alignments, we can create new annotations for a given target language, yielding high-quality annotated data for the target language, ready to use in an evaluation, without requiring more human annotation.
We focus on annotations which do not require token alignments (e.g., number of referential noun phrases that appear in a verse), as this avoids a source of noise (poor alignments) in annotation projection.

In this work, we describe our methods for creating the PrOnto dataset, and also provide experimental results demonstrating its utility as an evaluation resource.
We summarize our contributions as follows:

\begin{itemize}
    \item We publish evaluation datasets for 5 tasks across \numtrans{} New Testament translations in \numlangs{} languages.\footnote{These datasets and all of our code are available at \CodeRepo{}}
    \item We publish the system we used to create this dataset, which can be used by anyone to extend this dataset to any language that has a New Testament translation or a part of one.
    \item We perform experiments covering a wide range of languages with respect to typological variables and data-richness which demonstrate the utility of this dataset for assessing pretrained language model quality.
\end{itemize}

\section{Related Work}
Beginning with the publication of the first modern TLM, BERT \citep{devlin_bert_2019}, pretrained TLMs have had their quality assessed by applying them to a wide array of downstream tasks.
It is typical to apply the TLM in question to as many downstream evaluations as practically possible, since downstream tasks vary considerably in which properties of language they are sensitive to.
A syntactic parsing task, for example, is presumably more discriminative of formal aspects of grammar, while a sentiment analysis task is presumably more discriminative of meaning-related aspects of grammar.
All 11 of the tasks used to evaluate BERT in \citet{devlin_bert_2019} are meaning-oriented tasks, with natural language understanding (NLU) and question answering (QA) being heavily represented.

Most post-BERT English TLMs have followed its lead in favoring meaning-related tasks \citep[e.g.][\textit{inter alia}]{liu_roberta_2019,zhang-2022-improve}.
The English TLM evaluation dataset ecosystem has continued to grow, and some evaluation dataset suites have grown to encompass over 200 tasks \citep{bigbench}.
Among other high-resource languages, there is more variation: MacBERT \citep{cui-etal-2020-revisiting}, a Mandarin Chinese BERT, is evaluated using tasks comparable in kind and quantity to those used with BERT, while CamemBERT \citep{martin-etal-2020-camembert}, a French BERT, is evaluated with a large proportion of Universal Dependencies (UD) \citep{nivre_universal_2016} tasks.

The situation for low-resource languages is quite different.
Since annotated datasets are so rare and small for low-resource languages, most low-resource TLM evaluation has been centered on just a few datasets, all of which are fairly form-oriented in terms of what they are assessing models for.
Occasionally, a family of low-resource languages might have a high-quality evaluation dataset: for example, \citet{ogueji-etal-2021-small} train a low-resource TLM for 11 African languages, and evaluate on named-entity recognition (NER) using the MasakhaNER dataset \citep{adelani-etal-2021-masakhaner}. 
However, more often, low-resource languages do not have resources like this.

Much recent work on low-resource TLMs \citep[][\textit{inter alia}]{chau-etal-2020-parsing,chau-smith-2021-specializing,muller-etal-2021-unseen,gessler-zeldes-2022-microbert} uses only two datasets.
The first is UD corpora, which consist of human-annotated syntactic trees and tags which can be used for form-related tasks such as part-of-speech tagging and syntactic dependency parsing.
The second is the WikiAnn \citep{pan-etal-2017-cross} dataset, an NER dataset that was automatically generated for 282 languages based on the structure of Wikipedia hyperlinks.
While evaluations that use both of these datasets have proven to be useful, the UD dataset and to a lesser extent the WikiAnn dataset are both more form- than meaning-based in terms of what they assess in models.
This could mean that many low-resource TLM evaluations are missing important dimensions of model quality that cannot be assessed well by existing evaluation datasets.

Annotation projection is a technique at least as old as \citet{yarowsky-ngai-2001-inducing}, where token alignments are used to project noun phrase boundaries and part-of-speech tags across languages.
Much similar work has been done for other annotation types---just a few examples of works in this literature include \citet{pado} (semantic roles), \citet{asgari-schutze-2017-past} (tense), and \citet{enghoff-etal-2018-low} (named entity recognition).
It is also worth noting that the idea of using a large collection of Bible data for NLP/CL is not a new idea \citep{mccarthy-etal-2020-johns}.

\section{OntoNotes}
\begin{figure}
    \centering
    \includegraphics[width=\linewidth]{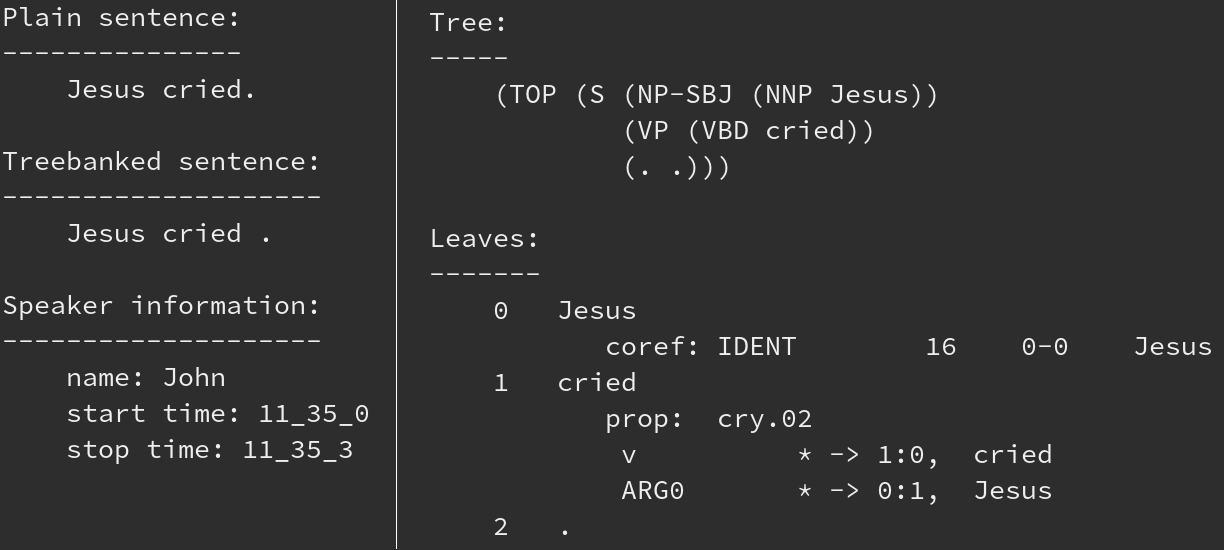}
    \caption{A sample verse, John~11:35, taken from OntoNotes. Note the annotations for tokenization, part-of-speech, constituency syntax, coreference, and argument structure. This file is in ``OntoNotes Normal Form'' (ONF), a human-readable format which OntoNotes provides its annotations in.}
    \label{fig:ontoexample}
\end{figure}
Before we describe our work, we briefly describe some important details of OntoNotes \citeplanguageresource{hovy_ontonotes_2006}.
OntoNotes is a multilayer annotated corpus whose English portion contains the Easy-to-Read Version (ERV) translation of the New Testament of the Christian Bible.
OntoNotes' major annotation types include coreference, Penn Treebank--style constituency syntax, NER, WordNet sense annotations, and PropBank argument structure annotations.
The ERV New Testament subcorpus of OntoNotes has all of these major annotation types with the notable exception of NER and WordNet sense annotation, which was not done for the New Testament.

An example annotation of John 11:35 is given in Figure \ref{fig:ontoexample}.
The ``Tree'' annotation has a Penn Treebank--style parse which includes an analysis of the sentence's syntactic structure as well as part-of-speech tags.
The ``Leaves'' section contains multiple annotation types which are anchored on the annotation's head token.
The \texttt{coref} type indicates a coreference annotation, which is then followed by coreference type, coreference chain ID, and token span information.
The annotation in Figure \ref{fig:ontoexample} tells us that: token 0, \textit{Jesus}, is the beginning of a new coreference mention; the coreference type of this mention is IDENT; the mention belongs to coreference chain 16; and this mention begins at token 0 and ends at token 0.

The \texttt{prop} type indicates the a PropBank annotation headed at the exponent of a predicate, typically a verb, and gives the PropBank sense ID of the predicate as well as the arguments of the predicate.
In the example in Figure \ref{fig:ontoexample}, the annotation tells us that: \texttt{cried} is the head of a PropBank predicate; the sense of the predicate is \texttt{cry.02}; the beginning of the \texttt{v} argument is headed at token 1, and its corresponding constituent is 0 levels up in the parse tree; and the beginning of the \texttt{ARG0} argument is headed at token 0 and its corresponding constituent is 1 level up in the parse tree.

For full details, we refer readers to the official documentation at \url{https://catalog.ldc.upenn.edu/docs/LDC2013T19/OntoNotes-Release-5.0.pdf}.

\section{Methods}
We would like to have more evaluation datasets for low-resource TLM evaluation, though constructing these for each individual language is expensive, as the creation of new datasets generally requires human annotation of some kind.
However, in this work, we propose a method for creating evaluation datasets without requiring additional human annotation.
New Testament translations are also highly common for low-resource languages because of missionary work, and OntoNotes' New Testament subcorpus is richly annotated.
Because the New Testament is partitioned into verses that are highly consistent across translations, it is possible to view verse boundaries as sentence-like alignments across translations, which would allow the projection of sentence-level annotations from OntoNotes to another New Testament translation.

This is the approach we take up: we propose five annotation projection methods, apply them to Bible translations, and perform evaluations to assess their utility.
More specifically, our goal is to take a New Testament translation in a \textit{target language}, align its verses with the verses present in OntoNotes, and then use OntoNotes' annotations to annotate the target language's translation, verse by verse.
Here, we describe the steps we take to process the data.

\subsection{Bible Translations}
We use all permissively-licensed New Testament translations available at \url{ebible.org}, a repository of Bible data, processing the proprietary XML format of these translations into our simple TSV format.
Some translations are very small or do not contain any of the New Testament, and we discard any with fewer than 500 verses overlapping with OntoNotes, which we do not count in our totals.
The final \numtrans{} translations cover a total of \numlangs{} languages.

It is important to note that there are many reasons to expect that Bible translations would be quite divergent from ideal naturalistic language data, such transcriptions of spontaneous conversation or formal oral narratives.
There are many common reasons which could produce expect this divergence, including: a translator's desire (for theological reasons or other) to use non-idiomatic expressions\footnote{An example of this in English is in Exodus 3:14, in which God's utterance is often rendered in English as ``I am that I am'', which is, in the author's opinion as a native English speaker, not idiomatic English due to \textit{that}'s inability to serve as the head of a free relative clause in this context. Presumably, the translator deliberately chose an unnatural English expression in order to attempt to preserve some grammatical properties of the original Hebrew.}; a translator's imperfect grasp of a target language; and the narrow distribution of the kinds of events that make up the subject matter of the Bible, to name just a few.
This is all to say that Bible data is distributionally quite different from more typical sources of language data.
While we have no choice but to use it in many low-resource situations, we must remember that for a given language, results on Bible data may not fully generalize to other kinds of data.

We additionally note that the ERV has particular deficiencies as an individual Bible translation.
The ERV's goal is to minimize the degree of reading comprehension needed in order for an individual to be able to read it, but in doing so, it sometimes engages in practices that are counter to our goals.
Most notably, the ERV is much less literal than many Bible translations, with sometimes entire clauses either being added or removed relative to the source text.
Additionally, the ERV sometimes combines verses (e.g.~Acts 1:16 and 1:17 are combined into a single verse in the ERV, 16--17, with no further indication as to which content belonged to which original verse), hindering projection.
Unfortunately, there is nothing to be done about these issues, as the ERV is what the creators of OntoNotes chose to use.

\subsection{Alignment}
\begin{figure}
    \includegraphics[width=\linewidth]{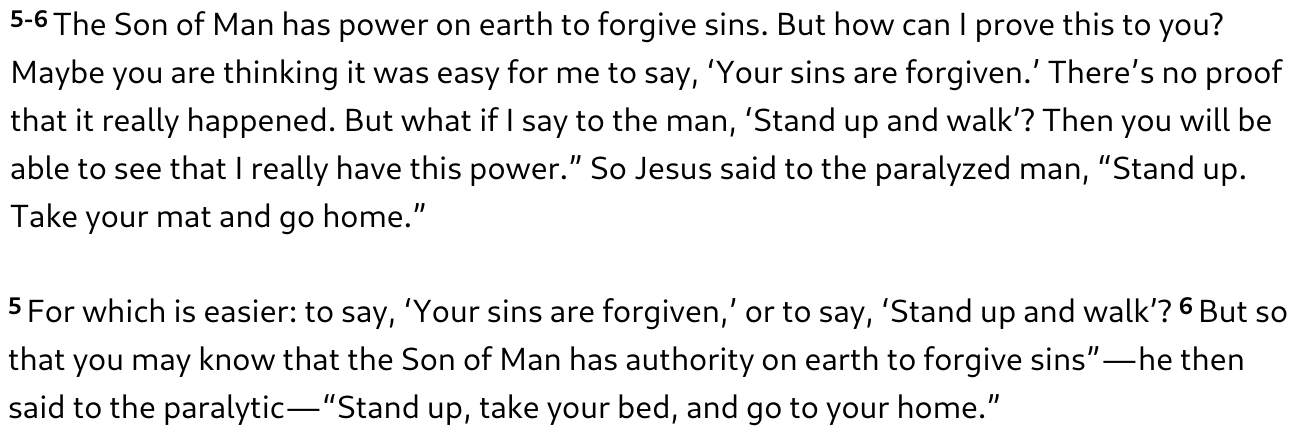}
    \caption{Matthew 9:5-6, as translated by the ERV (above) and the NRSVUE (below). In the ERV translation, verses 5 and 6 are fused, which means that no boundary between the two is indicated, and that their contents have been altered in linear ordering.}
    \label{fig:fusedverses}
\end{figure}
We parse OntoNotes' ONF files, and we assume that the target translation is given in a simple TSV format where each row contains the textual content of the verse as well as the verse's book, chapter, and verse number.
In an ideal situation, an OntoNotes sentence would correspond to exactly one verse in both the ERV and the target translation, but this is not always the case.
These are the possible complications:
\begin{enumerate}
    \item A verse contains more than one OntoNotes sentence. Some verses simply contain more than one sentence.
    \item An OntoNotes sentence spans more than one ERV verse. Verse boundaries are not guaranteed to coincide with sentence boundaries, so sometimes a sentence will begin in one verse and end in another. In OntoNotes, a sentence never spans more than two verses.
    \item The verse in either the ERV or the target translation has been combined with one or more other verses. Bible translators sometimes choose to combine verses and in such cases do not provide internal boundaries for the verses that have been merged.
\end{enumerate}
For determining a mapping, (1) presents no problem---we simply associate multiple OntoNotes sentences with a single verse.
For (2), we associate the sentence with both verses, retaining the information that a sentence spanned a verse boundary.
(For all of the tasks described in this paper, we discard verses that have sentences that cross verse boundaries, but the alignments are still constructed and ready to use.)
For (3), if verses have been combined in either the ERV or the target translation, we simply remove the combined verses from consideration.
In the ERV, combined verses are very rare, accounting for well under 1\% of all verses.
In other translations, this figure is also quite small.

\subsection{Tasks}
Once alignment is complete, we are prepared to generate task data.
We propose five tasks, all of which are sequence classification tasks either on single sequences or on paired (\`a la BERT's next sentence prediction) sequences.
While we do not pursue this in our present work, we expect that it may also be possible to produce annotations for token-level tasks using high-quality automatically generated word alignments.%
\footnote{For reasons of space, we will not give a formal treatment of our methods here. However, we encourage interested readers to refer to the publicly available implementation of each at \url{https://github.com/lgessler/pronto/tree/master/pronto/tasks}.}

A fundamental assumption for our approach is that some linguistic properties a sentence might have ought to be \textit{similar enough} in all languages to yield projected annotations which are useful for model evaluation.
Of course, short of examining every last verse, we cannot know with certainty that just because, for example, an English sentence has declarative sentence mood, its Farsi translation would also have declarative sentence mood.
But we do have reason to believe that sentence mood ought to be fairly well preserved across translations, given that sentence mood is so highly associated with semantic-pragmatic rather than formal aspects of language \citep{portner_mood_2018}, and so we can have some justification in assuming that sentence mood ought to be the same between translation pairs.
At any rate, regardless of the justifiability of this assumption, we contend that if this assumption does hold for a certain annotation type, then we should see differential performance across pretrained TLMs, which we will examine in \S \ref{sec:eval}.

\paragraph{Task 1: Non-pronominal Mention Counting (NMC)}
Predict the number of non-pronominal \textit{mentions} in a verse.
The intuition for this task is that it ought to require a model to understand which spans in a sentence could co-refer, which requires knowledge of both form and meaning.
A mention is a span of tokens, often but not always a noun phrase, that has been annotated for coreference, according to the OntoNotes-specific coreference annotation guidelines.\footnote{\url{https://www.ldc.upenn.edu/sites/www.ldc.upenn.edu/files/english-coreference-guidelines.pdf}}

It is important to point out that some entity must be mentioned at least \textit{twice} in a document in order to be annotated: if an entity is only mentioned once, then the mention is not annotated.
This makes this task somewhat pathological, because models will only be getting verse-level (not document-level) context, and this ought to make it impossible to tell in many cases whether a given markable (some tokens that \textit{could} be a mention) genuinely is a mention.
This is unfortunate, but this is not necessarily fatal for the utility of this task.%
\footnote{An alternative would be to simply count non-pronominal noun phrases in the parse tree, but this is not perfect either: some noun phrases, such as in \textit{on} $[_\mathrm{NP}$ \textit{the other hand}$]$, are never referential, a fact which coreference annotations are sensitive to but syntactic annotations are not. Without further work, it is unclear which is practically better, and we choose to use the coreference-based approach in this work.}

\paragraph{Task 2: Proper Noun in Subject (PNS)}
Predict whether the subject of the first sentence in the verse contains a proper noun.
To determine whether the subject contains a proper noun, we attempt to find a constituent labeled NP-SBJ in the main clause, and if we succeed in finding exactly one, we consider it a positive instance if any of the tokens within it are tagged with ``NNP'' or ``NNPS''.
Note that this does not necessarily mean that the \textit{head} of the subject is a proper noun: \textit{scholars}/NNS \textit{from}/IN \textit{Burundi}/NNP would count as a positive instance by our criterion, despite the fact that a common noun heads it.

\paragraph{Task 3: Sentence Mood (SM)}
Predict whether the mood of the main clause of the first sentence is declarative, interrogative, or imperative.
In Penn Treebank parse trees, sentence mood is encoded in the label of the highest constituent: for example, \texttt{S} and \texttt{S-CLF} are defined as having declarative sentence mood, \texttt{S-IMP} is imperative, and \texttt{SQ}, \texttt{SBARQ}, and \texttt{SQ-CLF} are interrogative.
If the top constituent does not have a label that falls into any of these categories, which likely means it is a sentence fragment or some other unusual sentence type, we discard it.

\paragraph{Task 4: Same Sense (SS)}
Given two verses $v_1$ and $v_2$, and given further that $v_1$ contains at least one usage of the predicate identified by sense label $s$, predict whether $v_2$ also has a usage of sense label $s$.
Note that in our formulation of this task, the sense label $s$ is explicitly given as an input rather than left unexpressed because otherwise the model would need to look for whether \textit{any} sense-usages overlap across the two verses, which is likely too hard.
Pairs are sampled so that negative and positive instances are balanced.

This task is perhaps the most suspect of all of our five proposed tasks given the great diversity of distinctions that may or may not be made at the word sense level.
For example, for the English word \textit{go}, Bukiyip has at least three different lexical items, distinguished by vertical motion relative to the mover's position at the beginning of the going event: \textit{nato} `go up, ascend'; \textit{nab\textipa{@}h} `go down, descend'; and \textit{narih} `go around, go at a level grade'.
As such, we should expect that performance will likely be nowhere close to 90\% even on non-English high-resource languages, as the English sense labels will likely often reflect distinctions which are either unexpressed or not specific enough for the target language's sense-inventory.
Still, we expect that for any given language, \textit{some} sense labels will still be appropriate when projected, and if this is the case, then we expect that higher-quality models will be able to perform better than lower-quality ones.
 
\paragraph{Task 5: Same Argument Count (SAC)}
Given two verses $v_1$ and $v_2$ which both feature a usage of the predicate identified by sense label $s$, predict whether both usages of $s$ have the same number of arguments.
Pairs are sampled so that negative and positive instances are balanced.
We do not require that the verses have \textit{exactly} one usage of $s$, which we do in the interest of using as many distinct verses as possible, though this may be interesting to consider in future work.

\section{Evaluation}
In order to evaluate our dataset, we implement a simple sequence classification model and apply it to our tasks using a wide range of pretrained TLMs.
We evaluate a wide range of languages and models in order to get as much information as possible about the utility of our methods.
These include several low-resource languages, but we also include some high- and medium-resource languages in order to get additional perspective.

\subsection{Languages}
The only work we were able to locate in the literature on low-resource TLMs that both worked on a wide range of languages and made all of their pretrained TLMs publicly available is \citet{gessler-zeldes-2022-microbert}, and we therefore include all of the languages they studied in their work.
These include the the low-resource languages Wolof, Sahidic Coptic, Uyghur, and Ancient Greek.
(Gessler and Zeldes also published models for Maltese, but we were unable to locate a permissively-licensed Maltese Bible.)
These also include Tamil and Indonesian, two medium-resource languages.

We additionally consider the high-resource languages French and Japanese, which may be interesting to look at given that they are both high resource and typologically similar to and divergent from English, respectively.
Any differences that emerge between French and Japanese could be indicative of typological distance degrading the quality of our projected annotations.
Additionally, both of these languages have high-quality monolingual TLMs, and it would be interesting to examine if different patterns emerge in high-resource settings.

Finally, we include two different English translations. 
First, we include the original translation used in OntoNotes, the ERV, because it ought to give us an upper bound on projected annotation quality: ERV annotations projected to the same ERV verses ought to have the highest possible quality.
Second, we include the Noah Webster's revision of the King James Version.
The Webster Bible differs from the KJV only in that mechanical edits were made to replace archaic words and constructions, and we include it in order to see if relatively small differences across translations (same language, slightly different register) are enough to cause major differences in task performance, which would then indicate differences in projected annotation quality.

\subsection{Model Implementation}
\label{sec:eval}
We use HuggingFace's \citep{wolf_huggingfaces_2020} off-the-shelf AutoModelForSequenceClassification model.
This model takes a pretrained TLM and adds a sequence classification head (with pretrained weights, if available).
The architectural details of this head vary depending on which exact model a pretrained TLM is for (e.g.~BertModel or RobertaModel), but most major models, including BERT and RoBERTa, simply use one (BERT) or two (RoBERTa) linear transformations that are applied to the [CLS] (or equivalent) token.
The model is trained with a low learning rate for a small number of epochs before it is evaluated on a held-out test set for each task.

\paragraph{Hyperparamters} Specifically, we use the default parameters for the \texttt{transformers} package, version 4.28.1, for the Trainer class, with the following exceptions.
Learning rate is set to 2e-5, batch size is set to 16, training epochs is set to 10 except for SM in which case it is 20, and weight decay for AdamW is set to 0.01.

\paragraph{NMC Capping} 
For NMC, while we always provide the genuine number of non-pronominal mentions in our dataset, in our experiments, we cap the maximum number of mentions at 3, labeling any sentence with more than 3 mentions as if it only had 3.
This was done to make the task easier, as the number of sentences with more than 3 mentions is very low, and the model subsequently suffers while trying to learn how to count higher than three.

\paragraph{Sequence Packing for SS and SAC}
Recall that for the SS and SAC tasks, the inputs include not only two verses but also a sense label.
First, we pack the two verses into a single input sequence, obeying any model-specific rules about where to put special tokens.
In a BERT style model, for example, the sequence would look like [CLS] $v_1$ [SEP] $v_2$ [SEP].
There are many ways the sense label $s$ could be provided as an input, but we choose to provide the label as an extra token after the final token of the base sequence.
To do this, we extend the vocabulary $\mathcal{V}$ with $|\mathcal{S}|$ more entries, where $\mathcal{S}$ is the inventory of sense labels, so that the new vocabulary has size $|\mathcal{V}| + |\mathcal{S}|$.
Senses are individually assigned to the new entries, and each sense is put after the final token, e.g. [CLS] $v_1$ [SEP] $v_2$ [SEP] $s$.

\paragraph{Metrics}
We report accuracy on all tasks.
Other more specialized metrics might be more informative for some tasks where e.g. the task is a binary classification problem or the label distribution is highly imbalanced, but we find that accuracy alone is sufficient to support our findings here, and choose to work with it exclusively to simplify the discussion.

\subsection{List of Bibles}
Our complete list of Bibles for the evaluation is as follows.
We format them so that our own abbreviation for them comes first, the full title follows, and the code for ebible.org's page follows in parentheses (append this code to \url{ebible.org/details.php?id=}).
\begin{enumerate}
\setlength{\itemsep}{-0.2em}
    \item ERV: Easy-to-Read version (engerv)
    \item WBT: Webster Bible (engwebster)
    \item IND: Indonesian New Testament (ind)
    \item TAM: Tamil Indian Revised Bible (tam2017)
    \item FRA: French Free Holy Bible for the World (frasbl)
    \item JPN: New Japanese New Testament (jpn1965)
    \item GRC: Greek Majority Text New Testament (grcmt)
    \item COP: Coptic Sahidic New Testament (copshc)
    \item UIG: Uyghur Bible (uigara)
    \item WOL: Wolof Bible 2020 Revision (wolKYG)
\end{enumerate}

\subsection{List of Pretrained Models}
Our complete list of pretrained models from HuggingFace Hub for the evaluation is as follows.
Note that some abbreviations are repeated because language will disambiguate which one is meant.
The models beginning with \texttt{lgessler/microbert} are taken from \citet{gessler-zeldes-2022-microbert}, and the suffixes indicate whether pretraining took place with just MLM (\texttt{-m}) or the combination of MLM and part-of-speech tagging (\texttt{-mx}).
(We refer readers to their paper for further details.)
\begin{enumerate}
\setlength{\itemsep}{-0.2em}
    \item \texttt{\footnotesize bert-base-multilingual-cased}: mBERT
    \item \texttt{\footnotesize xlm-roberta-base}: XLM-R
    \item \texttt{\footnotesize bert-base-cased}: BERT
    \item \texttt{\footnotesize distilbert-base-cased}: DistilBERT
    \item \texttt{\footnotesize roberta-base}: RoBERTa
    \item \texttt{\footnotesize camembert-base}: BERT
    \item \texttt{\footnotesize cl-tohoku/bert-base-japanese}: BERT
    \item \texttt{\footnotesize l3cube-pune/tamil-bert}: BERT
    \item \texttt{\footnotesize cahya/bert-base-indonesian-522M}: BERT
    \item \texttt{\footnotesize lgessler/microbert-...-m}: $\upmu$BERT-M (where \texttt{...} is one of \texttt{\footnotesize wolof, ancient-greek, indonesian, coptic, uyghur, tamil})
    \item \texttt{\footnotesize lgessler/microbert-...-mx}: $\upmu$BERT-MX
\end{enumerate}

\subsection{Results}

\paragraph{English}
\begin{table}[t]
    \centering
    \footnotesize
    \begin{tabular}{l|rrrrr}
    Model & NMC & PNS & SM & SS & SAC \\\hline
    \textbf{ERV}       & 49.59 & 72.60 & 91.56 & 50.43 & 50.69 \\
    DistilBERT         & 71.93 & 99.07 & 99.72 & 94.48 & 61.34 \\
    BERT               & 71.12 & 99.23 & 100.00 & 97.75 & 63.25 \\
    RoBERTa            & 70.03 & 98.76 & 99.86 & 89.53 & 50.69 \\
    mBERT              & 67.30 & 99.23 & 99.86 & 96.11 & 61.04 \\
    XLM-R              & 69.35 & 99.07 & 100.00 & 49.57 & 50.69 \\\hline\hline
    \textbf{WBT}       & 49.73 & 70.43 & 90.87 & 50.53 & 50.78 \\
    DistilBERT         & 55.99 & 84.67 & 92.81 & 72.15 & 61.54 \\
    BERT               & 52.86 & 83.13 & 94.88 & 76.06 & 64.51 \\
    RoBERTa            & 55.31 & 82.04 & 91.15 & 57.11 & 50.78 \\
    mBERT              & 53.54 & 85.29 & 93.22 & 79.08 & 60.55 \\
    XLM-R              & 53.68 & 84.67 & 93.22 & 49.47 & 50.78 \\
    \end{tabular}
    \caption[PrOnto results, English]{Task accuracy for English by model and translation. ERV is the Easy-to-Read Version, WBT is the Webster Bible.}
    \label{tab:pronto_english}
\end{table}

Results for our two English datasets are given in Table \ref{tab:pronto_english}.
A majority-label baseline is given in the row labeled with the translation (ERV or WBT), and results with several common pretrained English models as well as two multilingual models are given.

Looking first at our ``control'' dataset, the projection from the ERV translation onto itself, we can see that overall our models perform well above the majority class baseline, indicating that all of our tasks are not intractable, at least in the most easy setting.
It's worth noting that the Sentence Mood task is very easy in this condition, with two models getting a perfect score.
The hardest task is Same Argument Count, with the best model performing only 13\% higher than the baseline.
A striking pattern with the sequence-pair tasks is that the RoBERTa-family models perform at chance in three out of four cases.
The only obvious reason why this might be is that the other, BERT-family models are pretrained with a sequence-pair task (next sentence prediction), while RoBERTa is not.
We set this matter aside for now and note that even very popular and generally high-quality models can have anomalous performance on some tasks.

Turning now to the other English translation, WBT, we see that performance is lower on the whole but remains discernably higher than the baseline in all cases.
It is worth noting that the variety of English used in WBT, a slightly modernized form of Early Modern English, is likely quite out of domain for all of our models, and in this sense, the WBT could be thought of as a few-shot setting.
A pattern similar to the one for the ERV emerges where the RoBERTa-family models fail to do anything meaningful for the Same Argument Count task.

Overall, the results are in line with what we would expect given other published results which have evaluated the quality of these five pretrained models.
The monolingual models almost always do best for ERV and in three out of five tasks for WBT (SS and PNS, where mBERT does best). 
Among the monolingual models, excepting the anomalous RoBERTa cases described above, BERT most often performs best, with DistilBERT doing best in only two cases, which accords with findings that DistilBERT's quality is usually slightly lower than BERT's \citep{sanh_distilbert_2020}.
In sum, these results on English corroborate our claim that our five tasks are well-posed, not pathologically difficult, and indicative of model quality, at least in English settings.

\paragraph{Medium-resource Languages}
\begin{table}[t]
    \centering
    \footnotesize
    \begin{tabular}{l|rrrrr}
    Model & NMC & PNS & SM & SS & SAC \\\hline
    \textbf{FRA}       & 49.86 & 76.78 & 89.76 & 50.40 & 51.14\\
    BERT               & 57.63 & 82.35 & 92.81 & 67.43 & 64.04 \\
    mBERT              & 56.27 & 84.83 & 92.67 & 77.43 & 64.88 \\
    XLM-R              & 57.49 & 84.21 & 92.95 & 49.60 & 51.14 \\\hline
    \textbf{JPN}       & 51.30 & 76.47 & 91.41 & 50.15 & 50.64\\
    BERT               & 58.25 & 89.63 & 94.04 & 73.52 & 62.46 \\
    mBERT              & 59.21 & 88.24 & 93.21 & 79.74 & 51.36 \\
    XLM-R              & 54.98 & 88.85 & 95.15 & 49.85 & 50.64 \\\hline
    \textbf{IND}       & 49.15 & 72.95 & 92.36 & 50.37 & 50.87\\
    BERT               & 54.40 & 87.92 & 92.80 & 69.25 & 62.79 \\
    $\upmu$BERT-M      & 54.12 & 88.24 & 94.09 & 61.46 & 62.28 \\
    $\upmu$BERT-MX     & 53.98 & 87.12 & 93.80 & 59.87 & 62.10 \\
    mBERT              & 51.28 & 87.44 & 94.52 & 72.08 & 64.35 \\
    XLM-R              & 55.40 & 86.63 & 92.36 & 50.37 & 49.13 \\\hline
    \textbf{TAM}       & 49.59 & 74.77 & 91.56 & 50.51 & 50.65\\
    BERT               & 54.90 & 86.84 & 92.53 & 49.49 & 50.65 \\
    $\upmu$BERT-M      & 53.13 & 81.27 & 91.70 & 62.33 & 62.34 \\
    $\upmu$BERT-MX     & 52.32 & 82.51 & 91.29 & 62.92 & 63.11 \\
    mBERT              & 55.45 & 85.29 & 92.39 & 70.32 & 64.24 \\
    XLM-R              & 55.86 & 85.14 & 91.56 & 50.51 & 50.65 \\
    \end{tabular}
    \caption[PrOnto results, medium-resource]{Task accuracy for ``medium-resource'' languages by language and translation.}
    \label{tab:pronto_mr}
\end{table}

We turn now to our ``medium-resource'' languages in Table \ref{tab:pronto_mr}: French and Japanese at the higher end, and Indonesian and Tamil at the lower end.
For all four languages, XLM-RoBERTa continues to struggle with sequence-pair classification tasks, performing essentially at chance for all languages.%

For French and Japanese, the monolingual BERT model's performance is typically a bit better than either of the multilingual models' performance, with one exception: for the same-sense (SS) task, mBERT performs significantly better than the monolingual model.
Thus the broad picture of performance is what we'd expect, though this one surprising result shows that our tasks are broad in what they assess models for.

For Indonesian and Tamil, the $\upmu$BERT models perform slightly worse on average than mBERT, in line with the results reported by \citet{gessler-zeldes-2022-microbert}. 
Compared to the full-size monolingual models, the $\upmu$BERT models also are slightly worse on average, save for SS and SAC for Tamil, where performance is at-chance for the monolingual BERT.

\paragraph{Low-resource Languages}
\begin{table}[t]
    \centering
    \footnotesize
    \begin{tabular}{l|rrrrr}
    Model & NMC & PNS & SM & SS & SAC \\\hline
    \textbf{GRC} & 50.41 & 76.32 & 90.73 & 50.40 & 50.87\\
    $\upmu$BERT-M        & 52.59 & 81.11 & 90.18 & 60.58 & 61.80 \\
    $\upmu$BERT-MX       & 56.81 & 81.42 & 91.56 & 60.95 & 61.71 \\
    mBERT              & 57.36 & 83.13 & 91.70 & 65.34 & 50.87 \\
    XLM-R              & 55.99 & 76.32 & 91.42 & 49.60 & 50.87 \\\hline
    \textbf{COP} & 48.98 & 75.50 & 89.75 & 50.35 & 51.24\\
    $\upmu$BERT-M        & 50.75 & 78.76 & 89.75 & 61.32 & 62.70 \\
    $\upmu$BERT-MX       & 53.34 & 80.78 & 91.55 & 61.30 & 61.58 \\
    mBERT              & 49.52 & 75.50 & 89.75 & 52.79 & 51.24 \\
    XLM-R              & 48.84 & 75.50 & 89.75 & 50.35 & 51.24 \\\hline
    \textbf{UIG} & 49.37 & 73.53 & 89.96 & 50.23 & 50.78\\
    $\upmu$BERT-M        & 49.37 & 81.30 & 89.96 & 60.65 & 61.78 \\
    $\upmu$BERT-MX       & 51.19 & 78.45 & 90.10 & 61.51 & 62.12 \\
    mBERT              & 51.46 & 80.35 & 91.23 & 62.73 & 50.78 \\
    XLM-R              & 54.53 & 84.94 & 92.93 & 49.77 & 50.78 \\\hline
    \textbf{WOL} & 51.47 & 77.72 & 90.36 & 50.44 & 50.45\\
    $\upmu$BERT-M        & 51.47 & 77.72 & 90.36 & 59.78 & 61.05 \\
    $\upmu$BERT-MX       & 59.24 & 79.90 & 90.36 & 63.08 & 63.46 \\
    mBERT              & 57.35 & 84.75 & 91.65 & 66.49 & 54.46 \\
    XLM-R              & 56.51 & 82.32 & 91.01 & 50.44 & 49.55 \\
    \end{tabular}
    \caption[PrOnto results, low-resource]{Task accuracy for low-resource languages by language and translation.}
    \label{tab:pronto_lr}
\end{table}
Results for low-resource languages are given in Table \ref{tab:pronto_lr}.
Something that distinguishes the low-resource languages from the medium-resource languages and English is that many models now perform no better than the majority baseline.
Many of the Wolof and Coptic models perform no better than the baseline, and fewer but still some of the Uyghur and Ancient Greek models do not outperform the baseline.
For the $\upmu$BERT models, we note that the frequency with which this happens seems connected to dataset size: the tokens used by the $\upmu$BERT developers for each language were approximately 500K for Wolof, 1M for Coptic, 2M for Uyghur, and 9M for Ancient Greek.
This demonstrates that some of our tasks are too hard to be solved at all by a model if it falls below a quality threshold, which can be seen as a desirable trait.

Differences between the best-performing model and the baseline can be very small in some cases, such as for Sentence Mood in most languages.
This may indicate that sentence mood annotation projection is inappropriate for some target languages, though the fact that models still do differentiate themselves in how able they are to do it demonstrates that some properties of the target language can at least be correlated with the sentence mood of a translation-equivalent English sentence.
The performance gain relative to the baseline remains quite high for the two sense-related tasks.

\paragraph{Quality Assessment}
In addition to our main experimental findings, we find in supplementary experiments that our projected annotations for tasks 1-3 have quality that exceeds what would be expected from a random baseline by a sizeable margin.
We refer interested readers to Appendix \ref{sec:addtl} for details.

\section{Conclusion}
We have presented PrOnto, a publicly available dataset of evaluation tasks for pretrained language models for \numtrans{} New Testament translations in \numlangs{}. 
Overall, our results show that our tasks remain meaningful even when projected to languages which are typologically very different from English, and also even when they are performed by models that were trained on very little data.
The fact that pretrained models distribute relative to each other in our tasks mostly in the same way that they do for established evaluation tasks constitutes evidence that these tasks are indeed indicative of model quality. 
Moreover, while our intent was primarily to develop this resource for low-resource languages, we have shown that it is able to serve medium- and high-resource languages as well.

In future work, we intend to continue developing additional tasks.
There is still much data that has not been fully used in the OntoNotes annotations, and some tasks (such as SAC) would likely benefit from refinement or reformulation.
We further invite interested readers to consider contributing a task, as our annotation projection pipeline has been structured to make tasks very easy to author.

Beyond language model evaluations, one reviewer of this work has also suggested that scores on PrOnto could be interpreted as a kind of typological distance metric.
Moving from the observation described above that the quality of the projected annotations will correlate with a language's typological distance from English, the reviewer further observed that each target language ought to have an upper bound on system performance due to the annotation projection errors.
This means that, if we supposed we had a perfect system, its performance would reveal the projection error rate in its task performance metrics, and in doing so, reveal something about a language's typological proximity to English.
Of course, systems are not always perfect: for any given language, each one may do much better or worse than another.
In order to realize this vision, then, one would need to devise a way of accounting for confounds such as systems' individual strengths.
Still, we join our reviewer in thinking this could be a promising thread to pursue, as it would provide a means for computing a quantitative heuristic measure of a language's typological similarity to English using only a Bible translation.

\section*{Acknowledgments}
We thank Amir Zeldes for originally suggesting the core idea in this work, and we further thank Nathan Schneider and members of the NERT lab for helpful feedback on a draft of this work.
We also thank the maintainers of ebible.org for hosting the open-access Bibles which were used in this work.
We finally thank our reviewers for their exceptionalyl helpful comments.

\nocite{*}
\section{Bibliographical References}\label{sec:reference}

\bibliographystyle{lrec-coling2024-natbib}
\bibliography{custom}

\section{Language Resource References}
\label{lr:ref}
\bibliographystylelanguageresource{lrec-coling2024-natbib}
\bibliographylanguageresource{resources}

\clearpage
\appendix

\section{Additional Evaluation}
\label{sec:addtl}
We complement our findings above with some additional evaluations in order to gain more perspective on the quality of the projected annotations.
We look in detail at a particular target language, Hindi---specifically, we use the Hindi Contemporary Version Bible\footnote{\url{https://ebible.org/details.php?id=hincv}}.
For tasks 1 and 2, we parse the Hindi using a pretrained Stanza \citep{qi-etal-2020-stanza} UD parser, and use the UD parses to construct annotations for tasks 1 and 2.
For tasks 3, 4, and 5, we manually inspect 50 annotations per task in order to assess whether annotation projection was successful, and if not, why it failed.

\subsection{Tasks 1 and 2: UD Parser Comparison}
Annotations for task 1 (Non-pronominal Mention Counting) and task 2 (Proper Noun in Subject) are both reconstructable from a UD parse.
We use Stanza's pretrained Hindi model and parse all verses in the Hindi Bible.
For NMC, we look at each token in a verse, and increment the mention count iff the token is tagged as either \texttt{PROPN} or \texttt{NOUN}; and it is not the case that the token is tagged as \texttt{NOUN} and its dependency relation is \texttt{compound}\footnote{If a token is tagged as \texttt{NOUN} and its dependency relation is \texttt{compound}, this means that it is a noun modifying a noun, as in the first word of the phrase \textit{noun compound}. These cases are not counted in order to maintain consistency with OntoNotes, which does not treat the modifier of a noun compound pair as a separate markable.}.
For PNS, we perform a breadth-first search from the root to find the first token labeled as \texttt{nsubj}, and label the instance as positive if either the root of \texttt{nsubj} or any of its descendants are tagged as \texttt{PROPN}. 
If there is no \texttt{nsubj}, we treat the instance as negative.\footnote{In our previous implementation, if we could not locate a subject, we discarded the verse. However, for our analysis here, we must have exactly the same set of verses that were used for PrOnto, which is why we instead label an instance as negative if we cannot find a subject.}
As in the prior evaluation, we cap the maximum mention count at 3 and treat any larger values as 3.

Once we have constructed the second set of annotations using the UD parses, we need some way to compare them to each other.
For both tasks, we use accuracy and another metric to compare the annotations.
For NMC, we use mean squared error as a measure of how different the mention counts are on average.
For PNS, we use Jaccard similarity, since it is a binary task.
Since both the UD- and OntoNotes-based annotations are automatically constructed, we can treat neither as ground truth, so we also compare the PrOnto annotations to a random baseline for both tasks.
For all metrics, we expect that the PrOnto- and UD-based annotations ought to have the highest similarities, beating both of the baselines.

\begin{table*}[]
    \parbox{.45\linewidth}{\centering
    \begin{tabular}{l|cc}
        \textbf{Pair (NMC)} & \textbf{Acc.} & \textbf{MSE} \\\hline
        PrOnto, UD       & 0.517 & 1.199 \\
        PrOnto, Random   & 0.252 & 1.652 \\
        UD, Random       & 0.246 & 1.801 \\     
    \end{tabular}
    }
    \hfill
    \parbox{.45\linewidth}{\centering
    \begin{tabular}{l|cc}
        \textbf{Pair (PNS)} & \textbf{Acc.} & \textbf{Jaccard} \\\hline
        PrOnto, UD       & 0.810 & 0.680 \\
        PrOnto, Random   & 0.499 & 0.336 \\
        UD, Random       & 0.499 & 0.332 \\ 
    \end{tabular}
    }
    \caption[UD analysis results for tasks 1 and 2]{UD analysis results for tasks 1 and 2. For task 1, we consider accuracy and mean squared error. For task 2, we consider accuracy and Jaccard similarity.}
    \label{tab:task12}
\end{table*}

Results are given in Table \ref{tab:task12}.
Looking first at NMC, we see that as predicted, the PrOnto and UD annotations have the greatest similarities.
The random baseline has a low similarity, as could be expected given that this task has 4 possible labels.
It is worth considering whether an MSE of 1.199 might be high.
To some degree this divergence between the UD-based annotations and PrOnto is expected given that, as discussed above, an important limitation of the PrOnto mention count is that referential phrases that do not participate in coreference (i.e.~are only mentioned once in a document) are not annotated in OntoNotes, and this presumably accounts for at least some of the divergence between these two annotation-sets. 
Still, we see that our two annotation methods outperform the baselines, yielding evidence of their quality despite the fact that they are automatically constructed.
Turning now to PNS, we see the same pattern as before but more strongly. 
The similarity between PrOnto and the UD-based annotations is much stronger than between PrOnto and the majority or random annotations, as measured by both metrics.

In sum, while it is not possible from this analysis to determine the true annotation quality of either set of annotations (or indeed, even which one might be better), the fact that they both outperform a random baseline by a large margin shows that they at least agree on many cases.
While of course there is no guarantee that if an annotation is agreed upon by two different sources it is more likely to be true, it would be surprising if that were not more true than not in this situation.
We turn now to describe the remaining tasks (Sentence Mood, Same Sense, Same Argument Count), which cannot easily have their annotations constructed from UD parses.

\subsection{Task 3: Qualitative Evaluation}
First, for Sentence Mood (SM), we straightforwardly inspect the Hindi translation of the verse alongside the PrOnto annotation, and judge whether the existing label is correct.
(In order to make label judgments, we rely on the same criteria that were used in order to arrive at the labels in the PTB trees from which our SM labels were derived, as described above. These criteria, described in the PTB guidelines, are unproblematically applicable to Hindi.)
We compare our results from this procedure to a majority class baseline---recall that \textit{declarative} is by far the most common sentence mood, accounting for $\approx$90\% of all verses in the ERV.

We annotate a hundred Hindi instances using this procedure, and find that the PrOnto projected label is correct in 95 of 100 cases, while the majority class baseline is correct in 86 of 100 cases.
This constitutes evidence that while the projection process is not perfect, the annotation projection for this task is substantially better than guessing.

\subsection{Tasks 4 and 5: Qualitative Evaluation}
Tasks 4 and 5 (Same Sense, Same Argument Count) both have to do with meaning and argument structure.
In order to assess the PrOnto annotations for these tasks, we would like to know the following information:
\begin{enumerate}
    \item Do the two verses \textit{actually} contain usages of the senses in question?
    \item Do the findings in (1) violate foundational assumptions about either task? (For SAC, we always assume that both verses do contain a usage of the sense. For SS, we always assume that the first verse does contain a usage of the sense.) We consider an instance ``well-formed'' iff no foundational assumptions are violated.
    \item If we have found positively in (2), is the label for the task correct? (For SS, this means asking whether the label correctly identifies whether verse 2 has a genuine usage of the sense; for SAC, this means asking whether the label correctly identifies whether the two usages have the same argument count.)
\end{enumerate}

\begin{table*}[ht]
    \centering
    \begin{tabular}{l|ccc}
        \textbf{Task}           & \textbf{Total} & \textbf{Well-formed} & \textbf{Correct} \\\hline
        Same Sense (4)          &  50   &      36     &     30  \\
        Same Argument Count (5) &  50   &      34     &     29  \\ 
    \end{tabular}
    \caption[Analysis results for tasks 4 and 5]{\textbf{Analysis results for tasks 4 and 5}. An instance is ``well-formed'' if the assumptions about the data hold in the target language, and an instance is ``correct'' if it is well-formed and the projected annotation is correct.}
    \label{tab:task45}
\end{table*}

We note that making consistent decisions about (1) is very difficult: how can we precisely say whether one word in the English translation is ``the same'' as the word in the Hindi translation of a verse?
We propose the following procedure for determining this:
\begin{itemize}
    \item Given the English word that originally bore the PropBank annotation, attempt to identify a Hindi token that best captures its lexical meaning.
    \item Look up the candidate Hindi token in the dictionary of \citet{platts_1884}\footnote{Accessed at \url{https://dsal.uchicago.edu/dictionaries/platts/}}, and accept it as ``the same'' as the English word if and only if the same English word is listed in the relevant definition of the Hindi word.
\end{itemize}
This is not perfect, but it does give us a more reproducible way of making these decisions.
Note that these criteria do \textbf{not} require that the Hindi word have anything grammatically in common with the English word---part of speech, for example, may differ.
Relatedly, we note the difficulty of determining argument count in step (3), though here we can more easily rely on the PropBank frames for each PropBank lemma in telling us what arguments are possible.

We answer these questions for both tasks on a random sample of 50 instances for each task, each of which consists of two verses, a sense label, and the task annotation.
A baseline comparison is not possible for this task because a crucial part of the input for the task, the sense label, is obtained via projection and is not guessable by trivial means.

Our results are given in Table \ref{tab:task45}.
The picture that emerges for both tasks is similar: around 70\% of instances are well-formed, and around 60\% are correctly annotated.
(Remember that well-formedness is a precondition for correctness, so the correct instances form a subset of the well-formed instances.)
Higher would be better, of course, but given that these are numbers for annotation \textit{projection}, we take these numbers to be indicative of quite high quality for these two tasks, at least for the SS and SAC tasks.

Of the instances that were not well-formed, a couple of problems came up repeatedly.
First, English lemmas for highly frequent words (such as \textit{do} or \textit{be}) often participated in light verb constructions or other constructions with highly marginal verbal lexical content.
These were often realized in the Hindi translation as highly lexically contentful verbs, which led to ill-formedness.
Second, the English translation used in OntoNotes, the Easy-to-Read version, often uses expressions that diverge in content and literalness quite a bit compared to other translations. 
For example, compare Jude 1:23 in the ERV and NRSVUE translations:
\begin{itemize}
    \item NRSVUE: save others by snatching them out of the fire; and have mercy on still others with fear, hating even the tunic defiled by their bodies.
    \item ERV: Rescue those who are living in danger of hell’s fire. There are others you should treat with mercy, but be very careful that their filthy lives don’t rub off on you.
\end{itemize}
In the latter half of the verse, the ERV translators decided to be explicit about a thematic matter that the NRSVUE (and presumably the original Greek) leaves metaphorical.
The result is that the predicate \textit{rub} is introduced, which is present nowhere in the original Greek and is likely not present in other languages' translations given that \textit{rub off on} is an English idiom.
Compare this with a very literal English translation of the Hindi translation:
\begin{itemize}
    \item HINCVB: \textit{baakiyon ko aag mein se jhapatakar nikaal lo, daya karate hue saavadhaan raho, yahaan tak ki shareer ke dvaara kalankit vastron se bhee ghrna karo.} 
    \item HINCVB, translation: Dash in and snatch the remaining out of the fire, remain cautious while extending grace, to the extent that you hate even the clothes soiled by their bodies.
\end{itemize}
This instance involving \textit{rub} is representative of a handful of cases in which ill-formedness resulted from a creative translation. 
In this respect, we can see that the ERV is a Bible translation that is poorly suited to cross-lingual annotation projection.

It is interesting to consider the figures in Table \ref{tab:task45} against the performance of various models on these two tasks for high- and medium-resource non-English languages (cf.~Table \ref{tab:pronto_mr}).
At a glance, the median performance for these two tasks is somewhere in the low 60s for all languages, though it occasionally gets quite high (mBERT on Japanese scores 79.74\% for SS). 
Incidentally, we have also just seen here that label correctness for the Hindi--English language pair is around 60\%, at least according to our analysis methodology, which, given its rather strict criteria for word-equivalence in well-formedness, may be conservative. 
The evidence from this analysis thus gives us reason to believe that a ``good'' performance is probably not very much more than 60-70\% for most language pairs (since only around that many annotations are actually correct). 
If this is true, then when we also consider the distribution of scores in Table \ref{tab:pronto_mr}, we have strong reason to believe that the SS and SAC tasks are well-posed and useful for measuring model quality.

\end{document}